\documentclass[letterpaper, 10 pt, journal, twoside]{IEEEtran}

\usepackage[utf8]{inputenc}

\IEEEoverridecommandlockouts

\usepackage[american]{babel} %
\usepackage{cite}
\usepackage{float}
\usepackage{listings}
\usepackage{microtype}
\usepackage{booktabs}
\usepackage{url}
\usepackage{graphicx}
\usepackage[font=small,labelfont=bf]{caption}
\usepackage{subcaption}
\usepackage{gensymb}
\usepackage{array, multirow}
\usepackage{amssymb}
\usepackage{amsmath}
\usepackage{hyperref}
\usepackage{lipsum}
\usepackage{xcolor}
\usepackage{ifthen}
\usepackage{multicol}

\addto\extrasamerican{%

}
\usepackage{balance}

\DeclareMathOperator\erf{erf}

\newcommand{\azimuth}{\varphi_\text{az}}
\newcommand{\inclination}{\theta_\text{inc}}

\newcommand{\incidentangle}{\theta_0}
\newcommand{\reflectangle}{\theta_1}
\newcommand{\refractangle}{\theta_2}

\newboolean{showchanges}

\setboolean{showchanges}{false}

\newcommand{\added}[1]{\ifthenelse{\boolean{showchanges}}{\textcolor{blue}{#1}}{#1}}

\newcommand{\addedblock}[1]{\ifthenelse{\boolean{showchanges}}{{\color{blue}#1}}{#1}}
\newcommand{\addedthis}[0]{\ifthenelse{\boolean{showchanges}}{\color{blue}}{}}

\newcommand{\changed}[1]{\ifthenelse{\boolean{showchanges}}{\textcolor{orange}{#1}}{#1}}
\newcommand{\changedblock}[1]{\ifthenelse{\boolean{showchanges}}{{\color{orange}#1}}{#1}}

\makeatletter
\def\blfootnote{\gdef\@thefnmark{}\@footnotetext}
\makeatother

\begin{document}

\title{RadaRays: Real-time Simulation of Rotating FMCW Radar for Mobile Robotics via Hardware-accelerated Ray Tracing}
\author{Alexander Mock$^{1}$, Martin Magnusson$^{2}$, and Joachim Hertzberg$^{1}$%
\thanks{Manuscript received: July 11, 2024; Revised: October 24, 2024; Accepted January 1, 2025. This paper was recommended for publication by Editor A. Bera upon evaluation of the Associate Editor and Reviewers' comments.}%
\thanks{$^{1}$Alexander Mock and Joachim Hertzberg are with the KBS group of Osnabrück University, Germany. Joachim Hertzberg is additionally with the Plan-based Robot Control group of DFKI Niedersachsen, Germany; amock@uos.de, joachim.hertzberg@uos.de.}%
\thanks{$^{2}$Martin Magnusson with the RNP lab of the AASS research centre at Örebro University, Sweden; {\tt\small martin.magnusson@oru.se}}%
\thanks{The DFKI Niedersachsen (DFKI NI) is sponsored by the Ministry of Science and Culture of Lower Saxony and the VolkswagenStiftung}%
\thanks{Digital Object Identifier (DOI): see top of this page.}
}

\markboth{IEEE Robotics and Automation Letters. Preprint Version. Accepted January, 2025}
{Mock \MakeLowercase{\textit{et al.}}: RadaRays: Real-time Simulation of Rotating FMCW Radar}

\maketitle

\begin{abstract}
RadaRays allows for the accurate modeling and simulation of rotating FMCW radar sensors in complex environments, including the simulation of reflection, refraction, and scattering of radar waves.
Our software is able to handle large numbers of objects and materials \added{in real-time}, making it suitable for use in a variety of mobile robotics applications.
We demonstrate the effectiveness of RadaRays through a series of experiments and show that it can more accurately reproduce the behavior of FMCW radar sensors in a variety of environments, compared to the ray casting-based lidar-like simulations that are commonly used in simulators for autonomous driving such as CARLA.
Our experiments additionally serve as a valuable reference point for researchers to evaluate their own radar simulations.
By using RadaRays, developers can significantly reduce the time and cost associated with prototyping and testing FMCW radar-based algorithms.
We also provide a Gazebo plugin that makes our work accessible to the mobile robotics community.
\end{abstract}

\begin{IEEEkeywords}
Simulation and Animation, Range Sensing, Software Tools for Robot Programming, SLAM, Collision Avoidance
\end{IEEEkeywords}

\section{Introduction}
\IEEEPARstart{R}{adar} sensors have proven useful for mobile robotics, as they are robust to bad weather conditions or otherwise low-visibility air~\cite{mullen2000hyb, fritsche2016radar, bilik2023comp}.
Frequency modulated continuous wave (FMCW) radars are prevalent, as they offer high resolution and low power consumption.
Recently, well-known algorithms of mobile robotics such as odometry estimation~\cite{hong2020radarslam, burnett2021yeti, adolfsson2021cfear, burnett2021hero, aldera2022goes}, SLAM~\cite{malcom2019acomp, hong2020radarslam, adolfsson2022tbv, hilger2024randt}, place recognition~\cite{barnes2020under, suaftescu2020kidnapped, gadd2024radvlad}, and topometric localization~\cite{burnett2022we} have been adapted for radar measurements.
All of these algorithms %
provide passive, sensor data-observing state estimations of a robotic system.
For developing such algorithms, it is sufficient to use recorded sensor data~\cite{barnes2020oxford, kim2020mulran, sheeny2021radiate, burnett2023boreas, gadd2024oord}.
However, algorithms actively changing the robot state, such as navigation and exploration, cannot be developed just on such recordings.
Usually, they are developed either directly with the robot hardware or more conveniently on a robot simulator.
Realistic simulations for lidar are easy to access and are implemented in common robot simulators, such as CARLA~\cite{adosovitskiy2017carla}, Gazebo~\cite{melo2019simcomp}, and Isaac Sim~\cite{monteiro2019simulating}.
\changedblock{However, radar simulations require more sophisticated methods to reach the same level of realism, as radar is more noticeably different from the idealized ray model used in lidar simulations.
Radar simulation has to account for wider beams and stronger environmental influences such as absorptions and reflections~\cite{kamann2018auto}}.
Accordingly, there is a lack of radar simulations that are both realistic and computationally fast enough to be used in place of robotic hardware for developing new radar-based algorithms.

\begin{figure}[t]
    \vspace{0.25cm}
    \centering
    \includegraphics[trim={0 0 0 0},clip,width=0.9\linewidth]{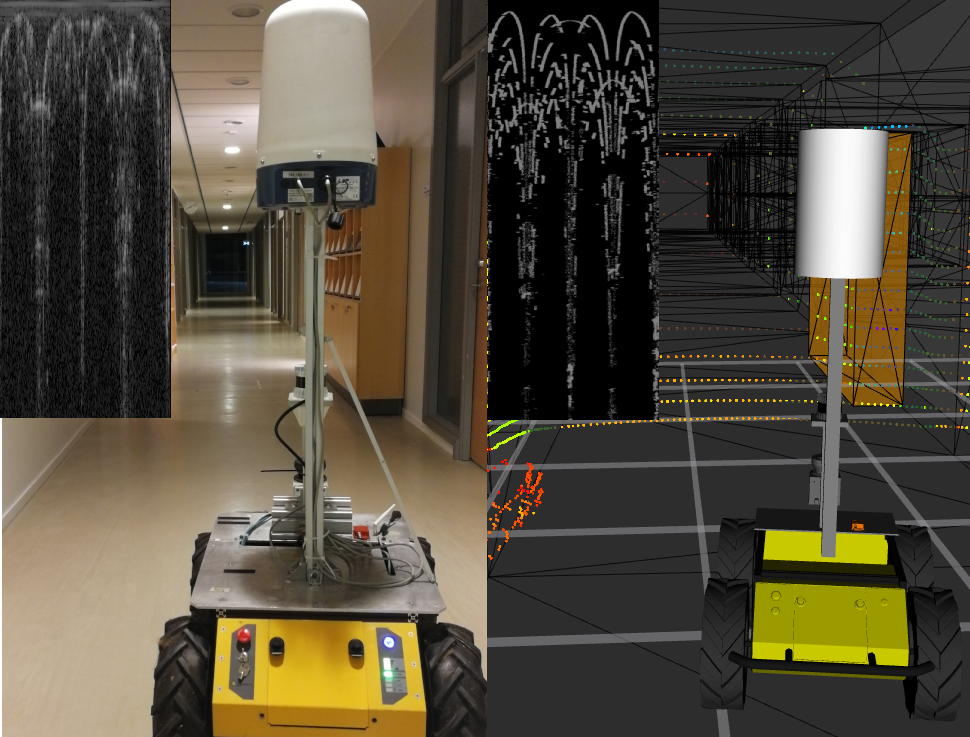}
    \caption{\textit{Left:} a Husky robot is scanning a corridor with a Navtech CIR radar producing the polar image on the top left. \textit{Right:} RadarRays simulation in a hand-modelled virtual scene. By applying material properties for radar wave reflections to every object we can closely replicate the real measurements. }
    \label{fig:oru_husky}
    \vspace{-0.2cm}
\end{figure}

We present an approach to overcome these existing limitations and propose RadaRays,\footnote{RadaRays is released under BSD 3-clause license and available at \\\url{https://github.com/uos/radarays}.} a real-time ray-tracing-based FMCW radar simulation method for mobile robotics.
One of its key benefits is its ability to handle complex and varied environments.
Our ray-tracing-based method can accurately simulate scenarios with many objects involving highly reflective or transmissive materials.
The method has the additional advantage of being highly efficient by leveraging the hardware acceleration of modern RTX GPUs.

Our main contribution is a simulation of rotating FMCW radars that balances high accuracy, supporting the development of SLAM algorithms, with sufficient speed to enable prototyping of novel navigation algorithms.
RadaRays comes with an easy-to-integrate Gazebo plugin.
Additionally, we introduce an experimental procedure and baseline to assess the performance of future radar simulators.

\section{Related work}

In 1978 Holtzman et al. published computer-aided simulations of synthetic-aperture radars (SAR)~\cite{holtzman1978}.
Their Point Scattering Model produced realistic and physics-based airborne radar imagery.
In 2007 Haynes et al. published the QuickBeam approach and software package, able to simulate meteorological radar data recorded from satellites.

In 2008, Scheiblhofer et al. proposed a method to simulate Frequency Modulated Continuous Wave (FMCW) radar systems~\cite{scheiblhofer2008}.
In 2010, Dudek et al. presented an approach for simulating radar wave propagation between the transmitter and receiver by ray tracing in addition to the circuit components~\cite{dudek2010}.
Both~\cite{scheiblhofer2008} and~\cite{dudek2010} focused on improving the technical designs of existing radars.
They provide realistic results, but none of them are real-time capable.

The autonomous driving simulator Carla~\cite{adosovitskiy2017carla} implements a radar sensor that initially casts rays along random directions to the scene using functionalities provided by Unreal Engine.
Once a ray hits an object in the scene, its attributes such as distance and relative speed are transferred to a partial sensor measurement.
However, neither multi-path reflections, nor the reflective and transmissive scattering properties of the materials by complete ray tracing are taken into account.

In 2017, Wheeler et al. published a deep learning-based approach to speed up computation times for realistic radar simulations~\cite{wheeler2017}.
They used a very simplified test scene that  suffices for testing radar systems installed in cars, but not for the application scenarios we are addressing.
In 2020, Tang et al. presented RSL-Net, a deep learning-based approach to localize rotating FMCW radars in satellite images~\cite{tang2020}.
The work involves generating synthetic radar images from satellite imagery.
It is unknown how realistic these simulations are, but they are at least realistic enough to solve the localization problem they addressed.
Weston et al. presented a deep-learned inverse sensor model to estimate occupancy probabilities of a grid map~\cite{weston2019}.
Two years later, they also proposed a deep-learned implicit model using adversarial optimization that was able to transform simulated elevation maps into realistic-looking radar images~\cite{weston2021}.
However, the approach is constrained to elevation information, precluding the consideration of different materials.
Additionally, the rate at which the simulations operate remains unspecified.

Overall, we consider ray tracing-based simulations~\cite{holtzman1978,dudek2010} as most promising for mobile robotics simulations compared to learned models due to their foundation in well-understood physical principles.
They require no large datasets for training, making them less data-dependent and immediately applicable.
Additionally, ray tracing provides deterministic and predictable outcomes, with direct control over simulation parameters, enhancing customization and versatility across different environments.
The drawbacks in run time were lately compensated by using the newest software and hardware for ray tracing~\cite{hirsenkorn2017ray, wald2019ray, holder2019, yilmaz2020radar, schuessler2021}.
They all speed up the ray tracing part by using the latest NVIDIA-RTX graphics cards and the NVIDIA-OptiX ray tracing engine~\cite{optix10}.
Schüßler et al. proposed a simulation for multiple input multiple output (MIMO) radars based on a shoot and bouncing rays algorithm~\cite{schuessler2021} capable of generating radar imagery of high realism for MIMO.
They were able to compute an array of 64 TX and 64 RX antennas in ca.\ 15 seconds.
With our approach, we adapt their work (for non-rotating radars from automotive) to the context of mobile robotics for rotating FMCW radars.
We focus on simulating in real-time even for high angular resolutions and providing a realistic simulation of the full spectral data of the radar return signals in the form of a polar intensity image.

\section{RadaRays}

A radar sensor emits a signal that is reflected, refracted and absorbed by the environment until it may return to a receiver.
By analyzing the returning signals, %
properties such as the locations or velocities of several objects can be estimated.
Translating this into a simulation requires precise formulations of the physical properties of radar waves propagating through the world.
First, consider the radar range equation:
\begin{equation}
    \label{eq:petot}
    P_e = \frac{P_s G^2 \lambda^2 \sigma}{\left(4 \pi \right)^3 R^4} \text{ .}
\end{equation}
It describes the signal strength $P_e$ that returns to the sensor after a radar signal of wavelength $\lambda$, strength $P_s$, and antenna gain $G$ is transmitted to a scene containing an object at range $R$ with a radar cross-section of $\sigma$.
While this equation explains the fundamental principles of radar wave propagation, it does not account for the complexities of realistic scenes.
We have built our radar simulation on these fundamental principles and tackled the challenges of complex scenes by employing ray tracing as our wave propagation technique.

\subsection{Ray Tracing} \label{sec:radarays:a}
\begin{figure}[b]
\centering
\begin{subfigure}{.6\linewidth}
    \centering
    \includegraphics[width=.98\linewidth]{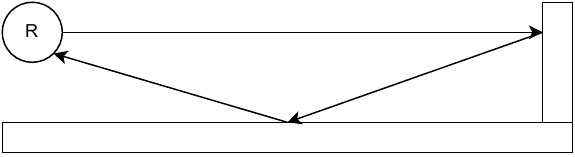}
    \includegraphics[width=.98\linewidth]{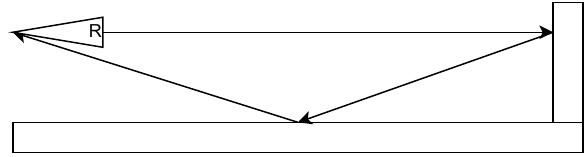}
    \caption{Uniform radar signal (top). Directed radar signal (bottom).}
    \label{fig:path_trace:uniform}
\end{subfigure}
\hspace{.01\linewidth}
\begin{subfigure}{.25\linewidth}
    \centering
    \includegraphics[trim={0 9cm 0 2cm},clip,width=0.98\linewidth]{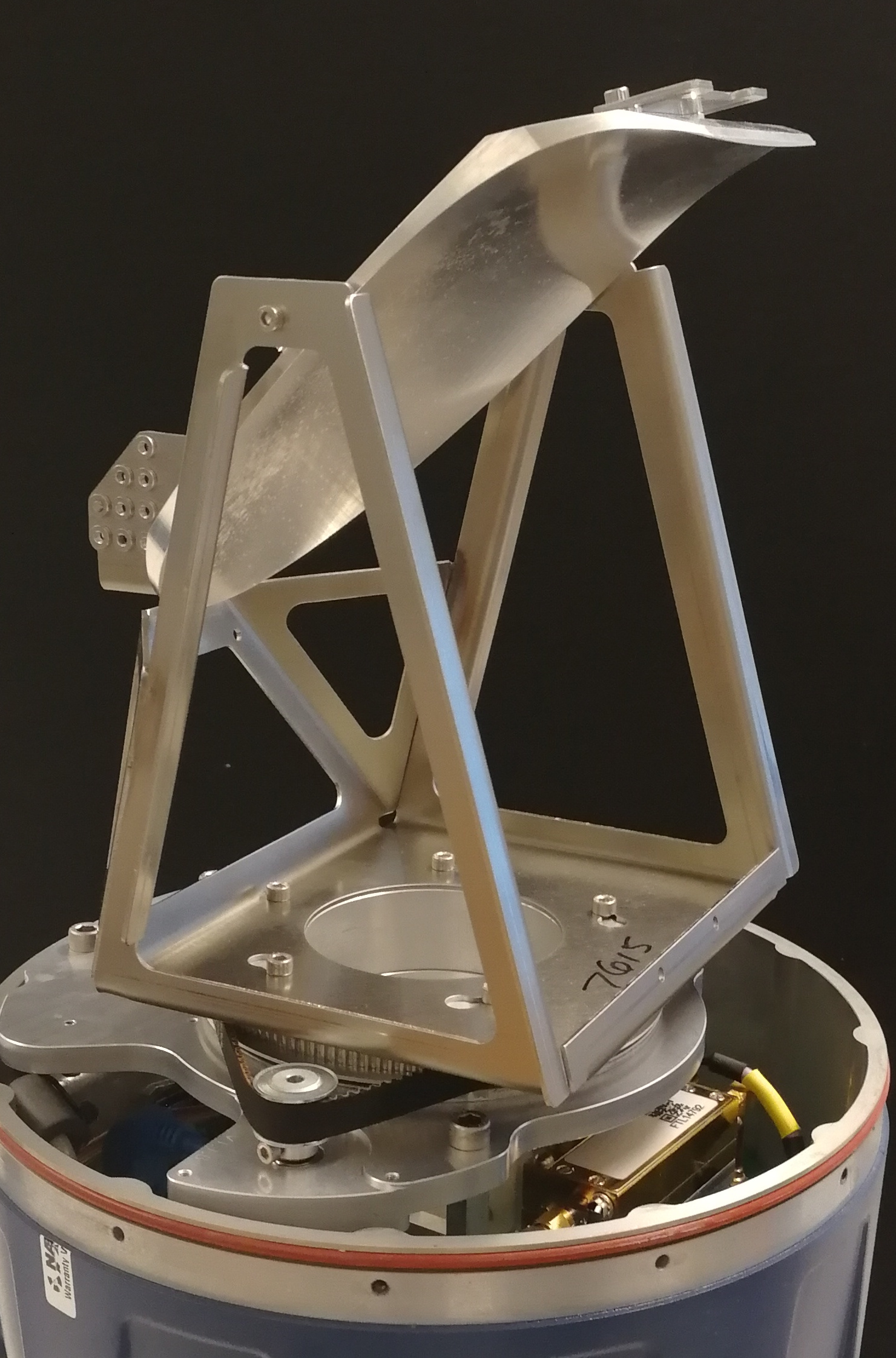}
    \caption{Radar mirror of Navtech CIR}
    \label{fig:path_trace:beam}
\end{subfigure}
\caption{Multi path reflections are important for classic radar as in (a) top. However, radar focusing mechanisms such as in (b) reduce the impact of these reflections as sketched in (a) bottom.}
\label{fig:path_trace}
\vspace{-0.2cm}
\end{figure}

We split up the emitted signal into ray samples, each of which represents a directed wave transporting a part of this signal's total energy.
The sum of each ray's energy again gives the total emitted signal energy preserving the law of conservation of energy.
If these rays are emitted uniformly in all directions, the energy is distributed over a spherical surface whose density automatically decreases with the distance to the power of four, according to \autoref{eq:petot}.
To model the signal emission process non-uniformly we let the user define a random density function $D$ from which the rays can initially be drawn.
With this function we can consider not only antenna gain $G$ but also any other focusing mechanisms in the simulated signal emission.
Such a focusing mechanism of the Navtech CIR sensor is shown in \autoref{fig:path_trace:beam}.

Our standard implementation includes four density functions $D_1$--$D_4$ that symmetrically sample rays in a cone shape.
The density functions can also be stretched along an axis to make the beam pattern more elliptic.
In the following we provide more details about the function $D_3$ as that is what we have used throughout our experiments.

First, we represent a ray by an azimuth and inclination angle $(\azimuth, \inclination)$.
Let the mean of the random distribution be the ray with direction $\overline{(\azimuth, \inclination)} = (0, 0)$.
To circularly draw around this mean, the representation in the random distributions is first transformed into the polar form: $\left(\azimuth, \inclination\right) = r_i \left(\cos(\omega), \sin(\omega)\right)$.
In this polar form, we can simply draw the angle $\omega$ from a uniform distribution $\mathcal{U} \in [-\pi,\pi)$.
What remains is the radius $r$, which describes the dispersion around the mean.
The following equation describes how we determine the radius for the drawn ray according to a probability $P$ to be inside the bounds given by the beam width.
\begin{equation}
    r = \frac{\mathcal{N}_0^1 \cdot b/2 }{\sqrt{2} \erf^{-1}\left(P\right)}
\end{equation}%
\textit{Notation:} $\mathcal{N}_0^1$: draw a random real number from a normal distribution with $\mu=0$ and $\sigma=1$.

\begin{figure}[b]
    \centering
    \includegraphics[trim={2.5cm 2.8cm 3cm 6cm},clip,width=0.7\linewidth]{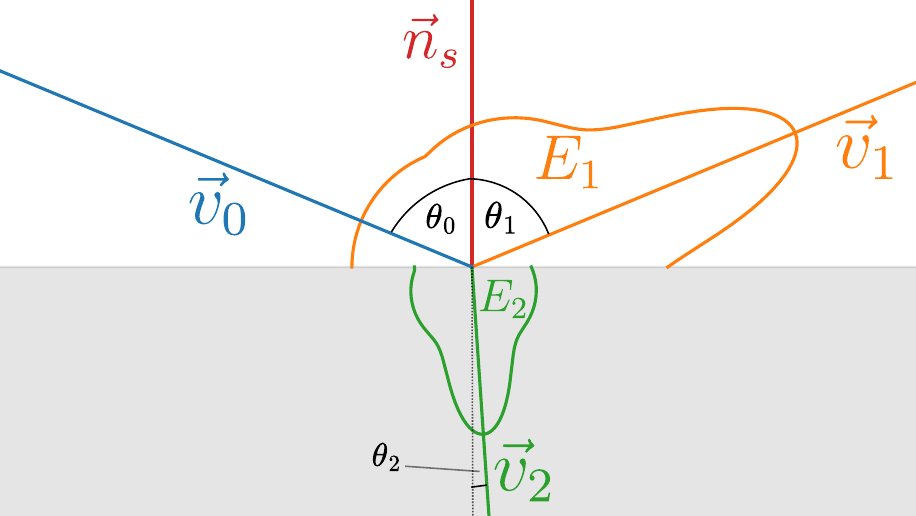}
    \caption{Standard reflection model of RadaRays. A blue incidence ray $\vec{v}_0$ passes from air medium (white) to another material (gray) that lets the wave travel with 0.4\,m/ns speed. The surface normal $\vec{n}_s$ is visualized in red. The directions of reflection $\vec{v}_1$, orange, and transmission $T$, green, are computed by Snell's law. The energy of transmission $E_2$ and reflection $E_1$ are computed by Fresnel equations. $E_1$ is the total area from the orange curve to the surface. The transmission energy $E_2$ is represented by the area between the green curve and the surface.}
    \label{fig:brdf}
\end{figure}

As the rays travel through the environment, they may be reflected, refracted, or absorbed, depending on the material properties of the objects.
Given the incident angle $\incidentangle$, we compute the mean reflection angle by $\incidentangle = \reflectangle$.
The ray part that is transmitted from medium 1 to medium 2 changes its velocity from $v_1$ to $v_2$.
We determine the mean angle of refraction $\refractangle$ by Snell's law $\sin(\incidentangle)/\sin(\refractangle) = v_1 / v_2$.
\changedblock{We then use the Fresnel equations} to determine how much total energy from the incident wave $E_0$ is reflected $E_1$ and how much energy goes into the other medium $E_2$.
After splitting $E_0$ up to $E_1$ and $E_2$ the law of conservation of energy must be satisfied with $E_0 = E_1 + E_2$.
For a planar surface, we can build two new rays and continue tracing the scene.
In reality, however, surfaces are mostly rough, causing the energy to reflect in many possible directions.
We model this phenomenon by a bidirectional energy density function of reflection $\dot{E}_1(\vec{v}_0, \vec{v}_1)$ which gives an energy density value for a given pair of incidence $\vec{v}_0$ and reflection ray $\vec{v}_1$.
To obey the law of energy conservation the integral $\int \dot{E}_1$ over the unit hemisphere on the outside of the surface must equal the total amount of reflected energy $E_1$.
Analogously, the integral $\int \dot{E}_2$ over the unit hemisphere on the inside of the surface must be equal to the refracted energy $E_2$.
Further, assuming the reflections to be isotropic in 3D gives us the property of the volume $E_1$ to be symmetric in the plane expressed by the normal vector $\vec{v}_0 \times \vec{n}_s$ with a peak at the mean reflection vector $\vec{v}_1$.
This lets us reduce the two inputs of the energy density function to just one; an angle difference $\omega$ between the mean reflection vector and another reflection vector $\theta_i$: $\omega = d(\reflectangle, \theta_i)$.
With $\omega \in [-\pi/2, +\pi/2]$ the law of energy conservation is satisfied by 
\begin{equation}
    E_1 = \int_{-\pi/2}^{+\pi/2} \dot{E}_1(\omega) d\omega \text{ .}
\end{equation}

This function has the same symmetric properties as $\dot{E}_1$ but has a peak in an angle difference of $\omega=0$ instead.
By integrating the energy density function $\dot{E}_1$ over an interval of the size of $b$ for any angular difference $\omega$ we can obtain the total reflected energy inside the beam:
\begin{equation}
    E_1(\omega) = \int_{\omega - b/2}^{\omega + b/2} \dot{E}_1(\alpha)d\alpha \text{ .}
\end{equation}
The actual shape of $\dot{E_1}$ can vary based on the material, incident angle, and reflection angle, and is represented in computer graphics using a bidirectional reflectance distribution function (BRDF).
RadaRays provides the capability to incorporate user-defined BRDFs for each material.
We include the following pre-implemented BRDF:
\begin{equation} \label{eq:brdf}
    E_1(\omega) = E_1 \left(A + B \cos(\omega) + S \cos(\omega)^{C}\right)
\end{equation}
which has three unknown parameters $A$, $B$, and $C$.
For these parameters the constraints $A \in [0,1]$, $B \in [0,1]$, $C \in \mathbb{R}$, and $A + B < 1.0$ must be satisfied.
We chose this approximation because it contains parts of commonly used BRDFs.
$A$ can be interpreted as the diffuse factor according to Lambertian reflection.
$B$ and $S$ are specular factors for mixing two glossy parts.
$C$ is the exponent for the second glossy part representing the specular hardness of the Phong reflection model~\cite{maier2018phong}.
The specular factor $S$ is given by $S = 1 - A - B$.
We chose this function since users who are familiar with CG shaders can guess them from experience, while the function is capable of approximating complex reflections as shown in \autoref{fig:brdf}, and remains computationally efficient.
If the user wants to consider more sophisticated reflection models such as subsurface scatting or microfacet modelling, user-defined BRDFs must be provided.
However, this may result in the loss of real-time simulation capabilities.
\addedblock{The preceding text describes the essential concepts of material modelling from the point of view of the RadaRays implementation.
The interested reader can find more information about BRDFs in general and the Fresnel equations, and how they relate to the general rendering equation, in \cite{cgpp-book}}.

The total energy function for the refractive part $E_2$ can be derived analogously.
Using $E_1$ and $E_2$, we first determine how much energy returns to the sensor.
For reflection, we considered two possible signal return paths: the signal is returning along the traveled path, or the signal is returning along the air path (multi-path reflection).
\changedblock{For the former, we assume that the transmitter and receiver are located at approximately the same point.}
This makes the angle difference between the mean reflection and returning ray equal to the difference between incidence and mean reflection angle.
For the latter, we determine the angle difference between the incidence ray and the direction from the intersection point back to the sensor.
However, once a radar has a mechanism to bundle the radiation to a beam, the impact of multi-path effects is reduced as shown in \autoref{fig:path_trace}.
If the incident ray is in another material, the same equations are used but with the refraction parts instead.
We collect all signals that return to the antenna in a list.
For the remaining signals, one new ray is spawned for reflection and one for refraction including their energy parts and emitted to the scene.
\changedblock{Each ray is traced until the maximum number of traces is exceeded or the energy carried by the ray is too low to be measured by the antenna.}

\subsection{Polar Image}

\begin{figure}[t]
    \centering
    \begin{subfigure}{.45\linewidth}
        \centering
        \includegraphics[trim={0 5cm 0 0},clip,width=0.98\linewidth]{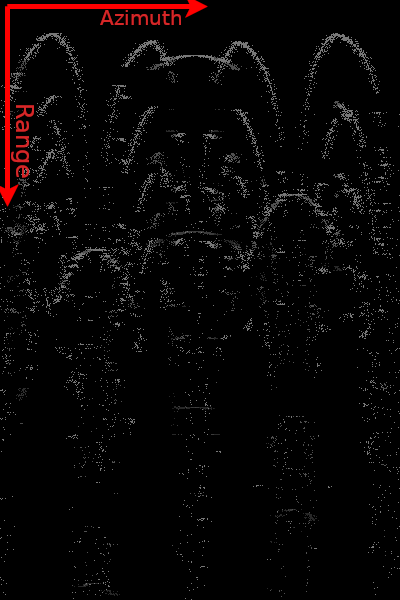}%
        \vspace{-0.1cm}
        \caption{Ray tracing only}
        \label{fig:radarays_images_a}
    \end{subfigure}
    \begin{subfigure}{.45\linewidth}
        \centering
        \includegraphics[trim={0 5cm 0 0},clip,width=0.98\linewidth]{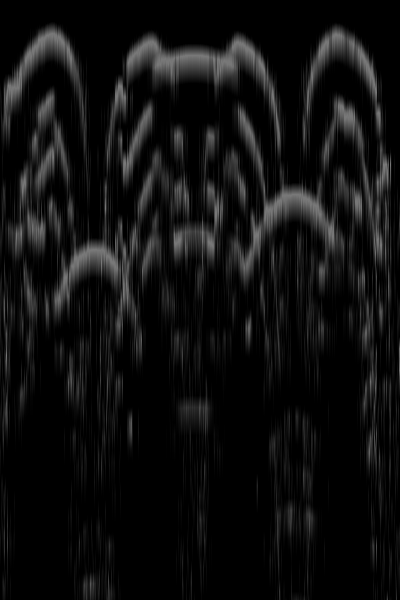}%
        \vspace{-0.1cm}
        \caption{+ system noise}
        \label{fig:radarays_images_b}
    \end{subfigure}\\
    \begin{subfigure}{.45\linewidth}
        \centering
        \includegraphics[trim={0 5cm 0 0},clip,width=0.98\linewidth]{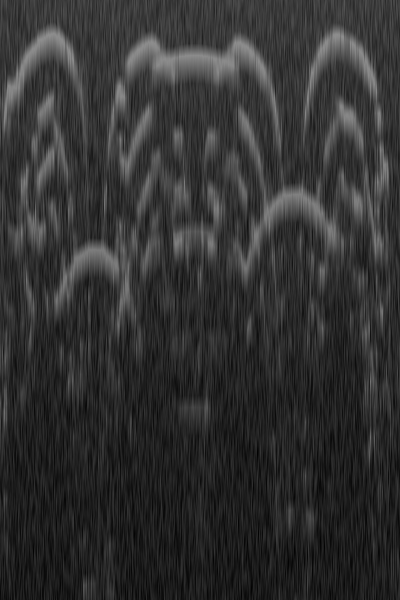}%
        \vspace{-0.1cm}
        \caption{+ ambient noise (full RadaRays model)}
        \label{fig:radarays_images_c}
    \end{subfigure}
    \begin{subfigure}{.45\linewidth}
        \centering
        \includegraphics[trim={0 5cm 0 0},clip,width=0.98\linewidth]{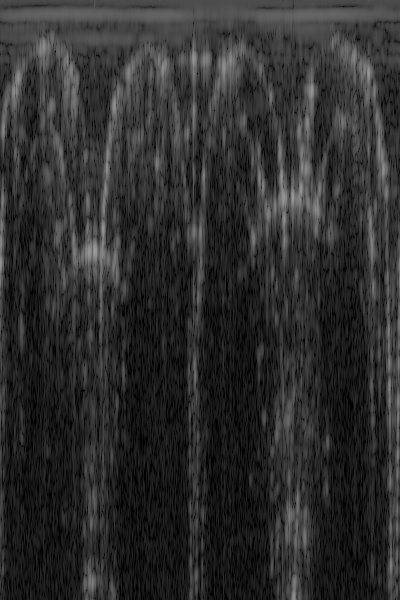}%
        \vspace{-0.1cm}
        \caption{Real radar image from a Navtech CIR radar}
        \label{fig:radarays_images_d}
    \end{subfigure}
    \caption{Three stages of RadaRays. a) The result after the ray tracing in Sec \ref{sec:radarays:a}. b) We apply post-processing noise to the perfect signals to model measurement noise. c) Perlin noise to model additional received signals that do not relate to a transmitted signal. d) The real Navtech CIR radar image recorded from the same position.} %
    \label{fig:radarays_images}
    \vspace{-0.2cm}
\end{figure}

Computing the whole procedure fills a list with returning signals for one intermediate azimuth angle of the rotating system.
We repeat this for each azimuth angle of a rotating radar system, thereby collecting the responses for each column in a polar image.
The columns and rows of a polar image represent discretized angles and distances.
This means that a pixel spans a rectangle over a certain angle and distance interval.
A simple lookup can be used to determine the corresponding pixel for each return signal.
Starting at 0, all signal strengths that fall into a pixel are then totaled.
Finally, we can calculate the relative signal strength $L_{E}$ in decibels from the total emitted energy $E_{0}$ and received energy $E_{R}$: $L_{E} = 10 \log_{10}\left( E_{R}/E_{0} \right) dB$. With $L_{E} = 0$, if $E_{0} = E_{R}$ and $L_{E} = -\infty$, if $E_{R} = 0$. 
For the polar image, $L_{E}$ is mapped from the range $[0, -\infty]$ to the pixel's value range $[255, 0]$ using parameters defining the maximum pixel value and a cut-off threshold at 0.
The results of the simulated signals assembled to a polar image are visualized in \autoref{fig:radarays_images_a}, which closely resembles the general shapes recorded by the actual sensor in the real environment visualized in \autoref{fig:radarays_images_d}.

\subsection{Noise}

\changedblock{At this point, we have modeled radar signals moving through a scene without impedance and an antenna measuring only the emitter's signal.
In reality, however, other effects influence the signals, antenna, and signal processing.
We classify all these effects, which we do not model in our scene, as noise.
Our software provides implementations for simulating ambient, system, and measurement noise.
For ambient and system noise, we implemented uniform noise and a slower but more sophisticated Perlin noise model.
Noise in frequency measurements causes errors in frequency shift, leading to erroneous distance computations.
To simulate these effects, we apply a customizable Gaussian blur to returning signals.}

\subsection{Software Design}

RadaRays uses Rmagine, a library to flexibly build ray-tracing-based sensor simulations~\cite{mock2023rmagine}, for all of its ray tracing computations.
In the current state, Rmagine provides two computing backends on which the ray tracing can be computed, Embree~\cite{embree14} or OptiX.
In doing so, we also inherit the same run time properties; e.g. with a larger number of triangles the simulation time increases logarithmically.
Because of the complexity of our sensor model, we inherit from Rmagine's \textit{OnDnModel}, which is used to model arbitrarily located and directed sensor rays.
Then, for a certain azimuth angle, the responses of the environment are collected and stored in a buffer, i.e. one column of the radar image.
This buffer is stored at the device of computation: when the Embree backend is used the radar measurements are stored in a RAM buffer, whereas when the OptiX backend is used the radar measurements are stored in a CUDA buffer.
RadaRays is a pure C\texttt{++} \changedblock{library}, whereas our provided Gazebo plugin is a ROS package that integrates RadaRays into Gazebo.

\section{Experiments}

\changedblock{Our series of experiments aims to evaluate RadaRays in terms of simulation accuracy and computational efficiency, focusing on its potential to serve as a replacement for physical sensors during software development.
The first experiment examines the seamless integration of RadaRays into SLAM software development workflows.
The second experiment evaluates the similarity between simulated and real-world sensor data.
The third experiment assesses the runtime performance of the software across various computing devices, offering a detailed analysis of its computational demands.
The fourth experiment showcases RadaRays’ capability to aid navigation algorithm development without the need for a physical sensor.
Finally, we investigate two challenging scenarios for real-time radar simulations: thin, highly reflective objects and the real-to-sim gap in deep-learned object detection models. 

In all our experiments, we use a radar configuration with a fixed number of 400 intermediate azimuth angles.
Further, for each azimuth angle, we initially sample 50 rays from the $D_3$ distribution with $b = 10\degree$ and $P=90\%$, and limit the maximum number of wave reflections to 4.}

\subsection{SLAM}

In the first experiment, we investigate how well RadaRays serves as a substitute for a real sensor in the software development process of simultaneous localization and mapping (SLAM) software. 
For this purpose, we use the YETI~\cite{burnett2021yeti} and CFEAR~\cite{adolfsson2021cfear} odometry methods.
Both approaches get a sequence of radar measurements as input data and estimate a trajectory from it.
In the experiment, we now systematically replace the real radar measurements with virtual ones generated by RadaRays.

\begin{figure*}[t]
    \centering
    \scalebox{1.0}[0.9]{\includegraphics[width=0.95\textwidth]{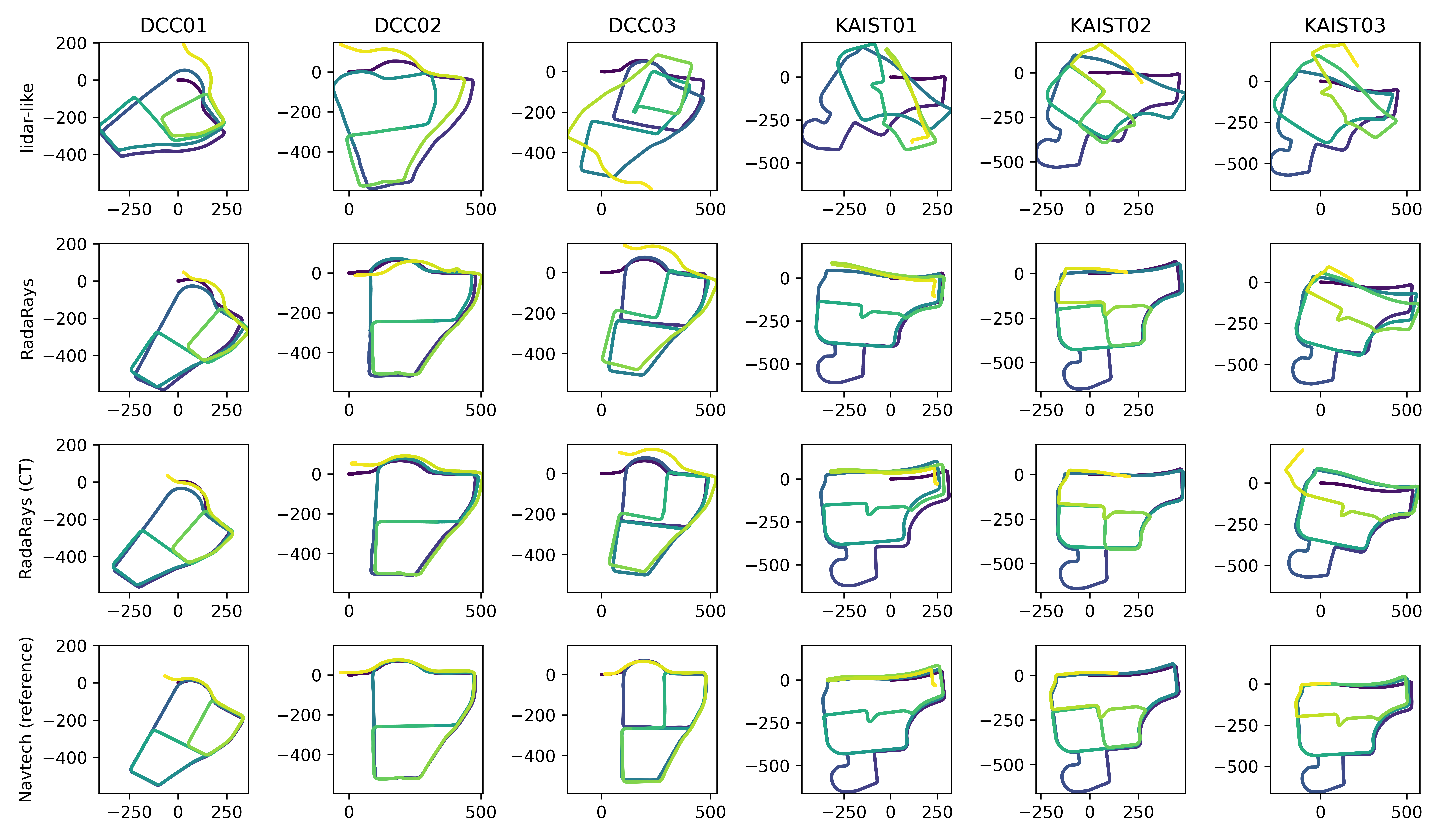}}
    \vspace{-0.3cm}
    \caption{Trajectories estimated by CFEAR for MulRan's DCC and KAIST sequences (columns) using simulated radar data as input (rows). From top to bottom: Lidar-like simulations, RadaRays using BRDF of \autoref{eq:brdf}, RadaRays using Cook-Torrance (CT) BRDF, actual Navtech data. Each trajectory is colored blue at the beginning and yellow at the end. All units are given in meters.}
    \label{fig:radarays_cfear2}
    \vspace{-0.25cm}
\end{figure*}

For the baseline radar data, we select the DCC and KAIST sequences from the MulRan datasets~\cite{kim2020mulran}.
Both were recorded in urban areas, whereas DCC additionally contains a long track along a forest.
For our experiments, we first reconstruct triangle meshes of these environments using LVR2~\cite{wiemann2018lvr}.
The resulting mesh of DCC contains 5,441,743 vertices, 8,845,352 triangles and covers an area of 719\,m $\times$ 534\,m;
the KAIST mesh contains 21,723,560 vertices, 12,408,626 triangles, covering 962\,m $\times$ 710\,m.
Then we localize the system in this mesh using MICP-L~\cite{mock2024micpl}.
Given the localization, we simulate radar images with various methods, always ensuring synchronization with the Navtech frames.
The first simulation method resembles the current CARLA implementation that considers only one reflection and uses narrow beams.
For this "lidar-like" simulator, we additionally disabled noise to ensure the best possible odometry estimation.
The second simulation method "RadaRays" employs the standard RadaRays reflection model.
The third method "RadaRays (CT)" uses a Cook-Torrance microfacet model~\cite{cook1982reflectance, walter2007microfacet} as user-defined BRDF for more realistic reflections.
As baseline we use the trajectories estimated on the actual Navtech data provided by the dataset.
For CFEAR, the trajectories for all simulation methods and the Navtech baseline is visualized in \autoref{fig:radarays_cfear2}.

We observe a high degree of alignment between the trajectories estimated from real data and those estimated from simulated data.
The relative trajectory errors (RTE) are listed in \autoref{tab:slam}.
We consider an odometry pipeline fails as soon as the relative trajectory error (RTE) exceeds 1\,m/m. 
This results in failure rates of RadaRays' standard reflection model of 0.22\,\% and 0.04\,\% for the DCC and KAIST sequences, respectively.
The overall average RTE for DCC is 5.94\,cm/m and for KAIST 4.94\,cm/m.

\begin{table}[t]
    \centering
    \caption{Using the lidar-like baseline and the RadaRays simulations with state-of-the-art radar odometry approaches YETI and CFEAR. We measure the performance by computing the relative trajectory errors (RTE) compared to the trajectory estimated on the real radar data provided by the MulRan sequences DCC and KAIST. ``--" indicates that the respective method is loosing track.}
    \label{tab:slam}
    \resizebox{\columnwidth}{!}{%
    \begin{tabular}{ c c c c c }
    \toprule
    & DS & lidar-like & RadaRays & RadaRays (CT) \\
    \midrule
    \multirow{6}{*}{\rotatebox[origin=c]{90}{YETI \cite{burnett2021yeti}}}
    & DCC01   & 6.04 cm/m & 4.83 cm/m &  \textbf{4.64 cm/m} \\
    & DCC02   &        -- & \textbf{8.97 cm/m} & 10.51 cm/m \\
    & DCC03   & 7.07 cm/m & 6.70 cm/m &  \textbf{6.60 cm/m} \\
    & KAIST01 &        -- & 6.38 cm/m &  \textbf{5.51 cm/m} \\
    & KAIST02 &        -- & 5.56 cm/m &  \textbf{5.06 cm/m} \\
    & KAIST03 &        -- & 5.52 cm/m &  \textbf{5.46 cm/m} \\
    \midrule
    \multirow{6}{*}{\rotatebox[origin=c]{90}{CFEAR \cite{adolfsson2021cfear}}}
    & DCC01   & 5.41 cm/m & \textbf{4.02 cm/m} & 4.23 cm/m \\
    & DCC02   & 9.56 cm/m & \textbf{8.60 cm/m} & 10.12 cm/m \\
    & DCC03   & 7.31 cm/m & \textbf{5.52 cm/m} & 5.87 cm/m \\
    & KAIST01 & 8.07 cm/m & 5.39 cm/m & \textbf{5.06 cm/m} \\
    & KAIST02 & 7.23 cm/m & 4.89 cm/m & \textbf{4.55 cm/m} \\
    & KAIST03 & 7.45 cm/m & \textbf{4.53 cm/m} & 4.73 cm/m \\
    \bottomrule
    \end{tabular}}
    \vspace{-0.1cm}
\end{table}

The used MulRan sequences encompass a wide range of diverse outdoor environments.
To evaluate scalability across different dataset providers, we select three sequences from the Oxford Radar dataset~\cite{barnes2020oxford} and generat a 3D mesh for each using PIN SLAM~\cite{pan2024pin}.
However, the resulting meshes contain artifacts that sometimes occlude the virtual scanner, leading to empty frames and causing the radar SLAM system to fail.
Additionally, the size of the generated meshes makes it impractical to manually remove the artifacts.
As an alternative, we identify the longest artifact-free trajectory in each sequence and recalculate the relative trajectory errors (RTEs).
For the sequences ``10-11", ``10-14", and ``11-12" the RTEs of RadaRays are measured as 7.89 cm/m, 8.58 cm/m, and 6.61 cm/m, respectively.

\subsection{Accuracy}

In addition to the application-specific experiments, we seek to provide a more general, image-based evaluation of RadaRays's simulations using the standard reflection model.
As perfectly accurate we define: Given a perfect model of the environment and a perfectly localized robot, RadaRays should generate measurements identical to those obtained from a real sensor.
Everything between identical and opposite is considered a degree of similarity.
We measure the degree of similarity for the DCC and KAIST sequences, applying a single material to the entire environment. 
For our standard reflection model~\autoref{eq:brdf}, we set the parameters to ($S=30$, $A=0.6$, $C=0.3$, $v=0.001$).
For the Cook-Torrance reflection model, we set the diffuse amount to $0.2$, the roughness to $0.5$, the refraction coefficient $n$ to $100$, and the transmittance to $0.1$.

In addition, we generate a dataset in Örebro University~(ORU): a robot with a Navtech radar driving along a corridor, as shown in~\autoref{fig:oru_husky}.
In contrast to DCC and KAIST, we handcraft the environment model, involving multiple materials with variable properties.
We use a training dataset with a stationary robot to search for the material parameters minimizing the Mutual Information Score (MIS)~\cite{igelbrink2019calib} using a broad grid search.
The results of each material including velocity of wave $v$ in [m/ns] and the three parameters of \autoref{eq:brdf}; $(v, A, S, C)$ are: Wall: $(0.001, 0.6, 0.3, 30.0)$, Wooden furniture: $(0.002, 0.6, 0.3, 70.0)$, Glass: $(0.05, 0.01, 0.04, 1900.0)$, Metal elements: $(0.0, 1.0, 1.0, 2000.0)$.
With these parameters, we reach a MIS of 0.215 on the test datasets.

To determine the similarity scores we then recorded a stationary dataset from another location and a dataset of the robot driving forth and back through the corridor.
The Structural Similarity Index (SSI)~\cite{whang2004ssi} and Mutual Information Score (MIS)~\cite{igelbrink2019calib} are used as similarity scores on time-synchronized real and simulated radar data.
The "lidar-like" simulation method described in the previous section serves as a baseline.
\autoref{tab:similarity} shows the results for the MulRan datasets, and for the optimized indoor model ORU.

\begin{table}[t]
    \centering
    \caption{Mean MIS and SSI in $\mu \pm \sigma$ of RadaRays' standard implementation compared to RadaRays' Cook-Torrance shader implementation (CT) using MulRan's KAIST and DCC sequences.
    Even though visual inspections give higher similarity to the CT version, the standard implementation achieves higher overall MIS and SSI scores.
    In addition, we measure MIS and SSI for our self-recorded ORU dataset which contains optimized materials.
    }
    \label{tab:similarity}
    \resizebox{\columnwidth}{!}{%
    \begin{tabular}{ c c c c c }
    \toprule
    & DS & lidar-like & RadaRays & RadaRays (CT) \\
    \midrule
    \multirow{3}{*}{\rotatebox[origin=c]{90}{MIS}}
    & DCC & 2.63E-4 $\pm$ 9.58E-5 & \textbf{5.05E-2} $\pm$ 1.90E-2 & 4.16E-2 $\pm$ 1.44E-2 \\
    & KAIST & 2.36E-4 $\pm$ 7.27E-5 & \textbf{3.92E-2} $\pm$ 1.12E-2 & 3.75E-2 $\pm$ 0.92E-2 \\
    & ORU & 1.35E-3 $\pm$ 7.79E-5 & \textbf{1.66E-1} $\pm$ 1.05E-2 & - \\
    \midrule
    \multirow{3}{*}{\rotatebox[origin=c]{90}{SSI}}
    & DCC & 1.61E-1 $\pm$ 3.90E-2 & \textbf{4.24E-1} $\pm$ 3.70E-2 & 3.89E-1 $\pm$ 4.04E-2 \\
    & KAIST & 1.78E-1 $\pm$ 2.28E-2 & \textbf{4.35E-1} $\pm$ 2.68E-2 & 4.05E-1 $\pm$ 2.90E-2 \\
    & ORU & 1.03E-2 $\pm$ 2.44E-4 & \textbf{2.89E-1} $\pm$ 3.79E-3 & - \\
    \bottomrule
    \end{tabular}}
    \vspace{-0.2cm}
\end{table}

Due to the high number of recorded samples, all error distributions are found to be pairwise statistically different, with p-values lower than 1E-8.
We observe the MIS is on average about 160.2 times higher for RadaRays than for the lidar-like baseline, whereas the SSI is on average 11.1 times higher.
RadaRays' MIS of the ORU datasets is 3.3 times higher than DCC, and 4.2 times higher than KAIST.

Overall, the results show that applying more sophisticated materials improves realism in terms of MIS and SSI.
This experiment also serves as a valuable reference for researchers to compare their radar simulations.
It offers a benchmark to evaluate and validate radar simulation techniques using publicly available MulRan datasets~\cite{kim2020mulran}.

\begin{figure*}[t]
    \centering
    \scalebox{1.0}[1.0]{\includegraphics[width=0.99\textwidth,trim={0 1cm 0 0.1cm},clip]{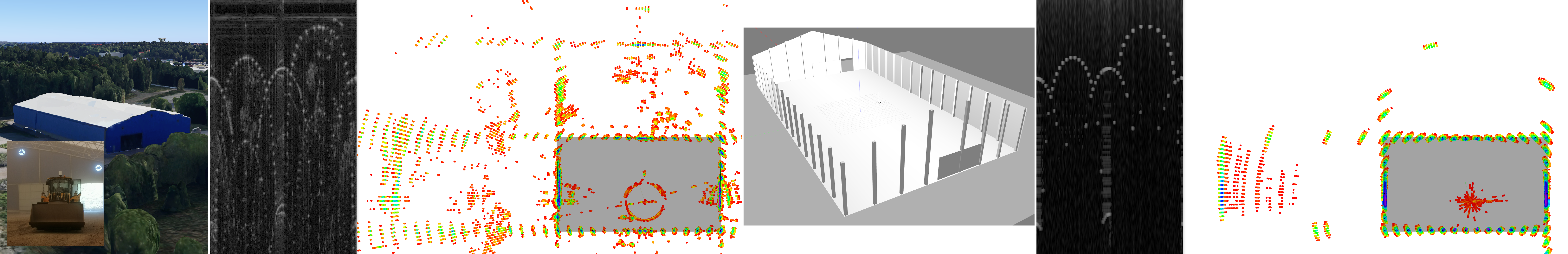}}
    \caption{\textit{Left images:} real world. From left to right: Google Earth view of the reflective hall, a wheel loader operating in it with a Navtech CIR radar mounted on top. A recorded polar image. Radar data is visualized in Euclidean space; the gray rectangle is the building captured top down. \textit{Right images:} RadaRays replicate. From left to right: Gazebo simulation, the simulated radar image, radar data visualized in Euclidean space. This experiment shows the limitation of RadaRays for highly reflective materials in combination with thin geometries.}
    \label{fig:volvo_ce}
    \vspace{-0.2cm}
\end{figure*}

\begin{table}[t]
    \centering
    \caption{Total run times (in ms) of computing a single radar image w.r.t. the number of initially cast rays $N_s$ per azimuth angle.}
    \label{tab:runtime_tot}
    \begin{tabular}{ c c c c c c c c }
    \toprule
     & Unit & 200 & 400 & 600 & 800 & 1000 \\
    \midrule
    \multirow{2}{*}{\rotatebox[origin=c]{90}{PC}}
    & Ryzen 7 3800X & 25.8 & 39.3 & 53.5 & 66.0 & 78.5 \\ 
    & RTX 2070 Super & 7.8 & 10.2 & 13.0 & 15.3 & 18.8 \\ 
    \midrule
    \multirow{2}{*}{\rotatebox[origin=c]{90}{NUC}}
    & i7 1165G7 & 50.9 & 79.1 & 106.9 & 134.1 & 161.5 \\ 
    & RTX 2060 Max-P & 10.6 & 14.1 & 18.0 & 21.8 & 26.0 \\ 
    \midrule
    \multirow{2}{*}{\rotatebox[origin=c]{90}{X1}}
    & i7 8750H & 97.5 & 145.2 & 173.8 & 205.4 & 239.4 \\ 
    & GTX 1050 Ti Max-Q & 71.4 & 113.9 & 165.4 & 188.1 & 264.1 \\
    \bottomrule
    \end{tabular}
    \vspace{-0.2cm}
\end{table}

\subsection{Computational Demands}

We provide a concise overview of viable hardware and software configurations compatible with RadaRays.
For our run time experiments, we choose the KAIST environment since its scale is realistic for mobile robot applications.
Using the standard RadaRays BRDF and the same parameters as in the previous experiments, we simulate one polar image for every radar pose of the KAIST03 sequence and measure the run time.
During our experiments we vary the following components to analyze their impact on the run times: the number of initially sampled rays per angle $N_s$, and the used implementation or computing unit.
The total number of initially cast rays can be determined by $N_r = 400 \cdot N_s$.
\autoref{tab:runtime_tot} shows the total run times of simulating one polar image on different devices.

The KAIST sequences contain both structured and unstructured parts, making it a suitable reference dataset for measuring the run time. 
As measured in \cite{mock2023rmagine}, the run time increases logarithmically with the map size.

\subsection{Navigation}

Next, we test the simulation for prototyping software for navigation.
As demonstration, we implement a simple fuzzy controller that lets a differential drive robot steer to the most free space in an office environment, using the complete information of the polar radar image we simulate using RadaRays.
The run time proved to be fast enough to enable reactive navigation, allowing the robot to successfully avoid obstacles of varying sizes.
In contrast, using the lidar-like baseline instead, the robot is colliding with the obstacles more often as the beams are missing small obstacles.
For visual support we refer to the accompanying video.

\subsection{Challenges in Real-Time Radar Simulation}

\subsubsection{Thin Highly-Reflective Geometries}

With a Navtech CIR radar mounted on a wheel loader, we record data inside a hall with challenging materials.
The hall’s walls are invisible by the radar, while the thin, metal pillars are highly reflective.
We hand-model the hall geometry, apply transparent materials to the walls and reflective materials to the pillars, and limit the ray tracing to two reflections.
The RadaRays simulation together with the real-world setup is shown in \autoref{fig:volvo_ce}.
It shows that with considerable effort in modeling, including finding properties for challenging materials, it is possible to achieve simulations that closely resemble real radar simulations.
Yet, some double reflections captured by the Navtech sensor are absent in the RadaRays simulations.
This discrepancy could stem from methodological constraints, as for this environment rays have a low chance of bouncing from one pillar to another.
Nonetheless, RadaRays provides more realistic data than lidar-like simulations, which fail to penetrate walls or simulate mirrored responses from pillars due to their lack of multi-reflection capabilities.

\subsubsection{Object Detection}

\changedblock{We test our approach for object detection suitability using the RADIATE dataset~\cite{sheeny2021radiate}, containing LiDAR, radar data, and vehicle annotations as ground truth for object detection tasks.
Using PIN SLAM~\cite{pan2024pin}, we generate meshes for three city sequences, spawn car meshes from dataset annotations, and simulate Cartesian radar images along PIN SLAM trajectories.
Testing the R-CNN from the RADIATE SDK, trained only on real radar data, shows it fails to detect most cars in the simulated images.

This failure may stem from two issues.
First, the approximations introduced to gain speed widen the real-to-sim gap so that the trained R-CNN becomes unable to generalize to our data.
Second, in this experiment we had to guess the reflection properties of the car's materials, as we could not extract them from the RADIATE dataset. %
Determining them by more sophisticated experiments could lead to more realistic results which might help with detecting cars using pre-trained object detectors. 
We are not aware of any other real-time simulator that is accurate enough so that object detectors trained on real data can transfer to simulated data.
Thus, the presented experimental setup can guide future work to evaluate more high-fidelity simulators.}

\section{Conclusion}

We have demonstrated that RadaRays, with its inclusion of material properties, noise, multi-path reflections, and large beam widths, offers a notable advantage over ray-casting-based lidar-like simulations as commonly used in simulators for autonomous driving such as CARLA.
Our analysis has revealed that the odometry estimated by YETI and CFEAR using simulated sensor data closely aligns with those estimated from real radar data, demonstrating the potential of RadaRays in replacing rotating FMCW radar sensors in the development of software for mobile robotics.
\changedblock{The consistent and substantial enhancements observed in image-based similarity measures (MIS and SSI) over the lidar-based baseline underline the efficacy of RadaRays in radar simulation.
While including a more sophisticated reflection model increases the simulation accuracy (MIS and SSI) it did not significantly increase the quality of odometry estimation over the default phenomenological model in our experiments.
Therefore, we believe our default implementation achieves a good balance between accuracy and run time efficiency for mobile robotics simulation applications.
The proposed evaluation strategy} together with RadaRays' unified interface to plug in user-defined BRDFs establishes a valuable framework for future advancements in radar simulation technology.
After choosing a ray tracing-based approach for a balanced trade-off between performance and realism, future work may focus on enhancing realism by implementing a full path tracer and investigating physical effects such as saturations and the occasional appearance of concentric rings.

\bibliographystyle{IEEEtran}
\bibliography{IEEEabrv,references}

\begin{thebibliography}{10}
\providecommand{\url}[1]{#1}
\csname url@rmstyle\endcsname
\providecommand{\newblock}{\relax}
\providecommand{\bibinfo}[2]{#2}
\providecommand\BIBentrySTDinterwordspacing{\spaceskip=0pt\relax}
\providecommand\BIBentryALTinterwordstretchfactor{4}
\providecommand\BIBentryALTinterwordspacing{\spaceskip=\fontdimen2\font plus
\BIBentryALTinterwordstretchfactor\fontdimen3\font minus
  \fontdimen4\font\relax}
\providecommand\BIBforeignlanguage[2]{{%
\expandafter\ifx\csname l@#1\endcsname\relax
\typeout{** WARNING: IEEEtran.bst: No hyphenation pattern has been}%
\typeout{** loaded for the language `#1'. Using the pattern for}%
\typeout{** the default language instead.}%
\else
\language=\csname l@#1\endcsname
\fi
#2}}

\bibitem{mullen2000hyb}
L.~Mullen and V.~Contarino, ``{Hybrid LIDAR-radar: Seeing through the
  scatter},'' \emph{{IEEE} Microwave}, vol.~1, no.~3, pp. 42--48, 2000.

\bibitem{fritsche2016radar}
P.~Fritsche, S.~Kueppers, G.~Briese, and B.~Wagner, ``{Radar and LiDAR
  Sensorfusion in Low Visibility Environments},'' in \emph{ICINCO (2)}, 2016,
  pp. 30--36.

\bibitem{bilik2023comp}
I.~Bilik, ``{Comparative Analysis of Radar and Lidar Technologies for
  Automotive Applications},'' \emph{IEEE Intell. Transp. Syst. Mag.}, vol.~15,
  no.~1, pp. 244--269, 2023.

\bibitem{hong2020radarslam}
Z.~Hong, Y.~Petillot, and S.~Wang, ``{RadarSLAM: Radar based Large-Scale SLAM
  in All Weathers},'' in \emph{Int. Conf. Intell. Robots Syst. (IROS)}.\hskip
  1em plus 0.5em minus 0.4em\relax IEEE, 2020.

\bibitem{burnett2021yeti}
K.~Burnett, A.~P. Schoellig, and T.~D. Barfoot, ``{Do We Need to Compensate for
  Motion Distortion and Doppler Effects in Spinning Radar Navigation?}''
  \emph{{IEEE} Robot. Automat. Lett.}, vol.~6, no.~2, pp. 771--778, 2021.

\bibitem{adolfsson2021cfear}
D.~Adolfsson, M.~Magnusson, A.~Alhashimi, A.~J. Lilienthal, and H.~Andreasson,
  ``{CFEAR Radarodometry - Conservative Filtering for Efficient and Accurate
  Radar Odometry},'' in \emph{Int. Conf. Intell. Robots Syst. (IROS)}.\hskip
  1em plus 0.5em minus 0.4em\relax IEEE, 2021, pp. 5462--5469.

\bibitem{burnett2021hero}
K.~Burnett, D.~J. Yoon, A.~P. Schoellig, and T.~D. Barfoot, ``{Radar Odometry
  Combining Probabilistic Estimation and Unsupervised Feature Learning},'' in
  \emph{Robotics: Science and Systems (RSS)}, 2021.

\bibitem{aldera2022goes}
R.~Aldera, M.~Gadd, D.~De~Martini, and P.~Newman, ``{What Goes Around:
  Leveraging a Constant-Curvature Motion Constraint in Radar Odometry},''
  \emph{{IEEE} Robot. Automat. Lett.}, vol.~7, no.~3, pp. 7865--7872, 2022.

\bibitem{malcom2019acomp}
M.~Mielle, M.~Magnusson, and A.~J. Lilienthal, ``{A comparative analysis of
  radar and lidar sensing for localization and mapping},'' in \emph{Eur. Conf.
  Mobile Robots (ECMR)}.\hskip 1em plus 0.5em minus 0.4em\relax IEEE, 2019, pp.
  1--6.

\bibitem{adolfsson2022tbv}
D.~Adolfsson, M.~Karlsson, V.~Kubelka, M.~Magnusson, and H.~Andreasson, ``{TBV
  Radar SLAM – Trust but Verify Loop Candidates},'' \emph{{IEEE} Robot.
  Automat. Lett.}, vol.~8, no.~6, pp. 3613--3620, 2023.

\bibitem{hilger2024randt}
M.~Hilger, N.~Mandischer, and B.~Corves, ``{RaNDT SLAM: Radar SLAM Based on
  Intensity-Augmented Normal Distributions Transform},'' in \emph{Int. Conf.
  Intell. Robots Syst. (IROS)}.\hskip 1em plus 0.5em minus 0.4em\relax IEEE,
  2024.

\bibitem{barnes2020under}
D.~Barnes and I.~Posner, ``{Under the Radar: Learning to Predict Robust
  Keypoints for Odometry Estimation and Metric Localisation in Radar},'' in
  \emph{Int. Conf. Robot. Automat. (ICRA)}.\hskip 1em plus 0.5em minus
  0.4em\relax IEEE, 2020, pp. 9484--9490.

\bibitem{suaftescu2020kidnapped}
{\c{S}}.~S{\u{a}}ftescu, M.~Gadd, D.~De~Martini, D.~Barnes, and P.~Newman,
  ``{Kidnapped radar: Topological radar localisation using
  rotationally-invariant metric learning},'' in \emph{Int. Conf. Robot.
  Automat. (ICRA)}.\hskip 1em plus 0.5em minus 0.4em\relax IEEE, 2020, pp.
  4358--4364.

\bibitem{gadd2024radvlad}
M.~Gadd and P.~Newman, ``Open-radvlad: Fast and robust radar place
  recognition,'' in \emph{2024 IEEE Radar Conf.}\hskip 1em plus 0.5em minus
  0.4em\relax IEEE, 2024, pp. 1--6.

\bibitem{burnett2022we}
K.~Burnett, Y.~Wu, D.~J. Yoon, A.~P. Schoellig, and T.~D. Barfoot, ``{Are We
  Ready for Radar to Replace Lidar in All-Weather Mapping and Localization?}''
  \emph{{IEEE} Robot. Automat. Lett.}, vol.~7, no.~4, pp. 10\,328--10\,335,
  2022.

\bibitem{barnes2020oxford}
D.~Barnes, M.~Gadd, P.~Murcutt, P.~Newman, and I.~Posner, ``{The Oxford Radar
  RobotCar Dataset: A Radar Extension to the Oxford RobotCar Dataset},'' in
  \emph{Int. Conf. Robot. Automat. (ICRA)}.\hskip 1em plus 0.5em minus
  0.4em\relax IEEE, 2020, pp. 6433--6438.

\bibitem{kim2020mulran}
G.~{Kim}, Y.~S. {Park}, Y.~{Cho}, J.~{Jeong}, and A.~{Kim}, ``{MulRan:
  Multimodal Range Dataset for Urban Place Recognition},'' in \emph{Int. Conf.
  Robot. Automat. (ICRA)}.\hskip 1em plus 0.5em minus 0.4em\relax IEEE, 2020,
  pp. 6246--6253.

\bibitem{sheeny2021radiate}
M.~Sheeny, E.~De~Pellegrin, S.~Mukherjee, A.~Ahrabian, S.~Wang, and A.~Wallace,
  ``{RADIATE: A Radar Dataset for Automotive Perception in Bad Weather},'' in
  \emph{Int. Conf. Robot. Automat. (ICRA)}.\hskip 1em plus 0.5em minus
  0.4em\relax IEEE, 2021, pp. 1--7.

\bibitem{burnett2023boreas}
K.~Keenan~Burnett, D.~J. Yoon, Y.~Wu, A.~Z. Li, H.~Zhang, S.~Lu, J.~Qian, W.-K.
  Tseng, A.~Lambert, K.~Y. Leung, A.~P. Schoellig, and T.~D. Barfoot,
  ``{Boreas: A multi-season autonomous driving dataset},'' \emph{Int. J.
  Robotics Research}, vol.~42, no. 1-2, pp. 33--42, 2023.

\bibitem{gadd2024oord}
M.~Gadd, D.~De~Martini, O.~Bartlett, P.~Murcutt, M.~Towlson, M.~Widojo,
  V.~Muşat, L.~Robinson, E.~Panagiotaki, G.~Pramatarov, M.~Alexander~Kühn,
  L.~Marchegiani, P.~Newman, and L.~Kunze, ``{OORD: The Oxford Offroad Radar
  Dataset},'' \emph{IEEE Transactions on Intelligent Transportation Systems},
  vol.~25, no.~11, pp. 18\,779--18\,790, 2024.

\bibitem{adosovitskiy2017carla}
A.~Dosovitskiy, G.~Ros, F.~Codevilla, A.~Lopez, and V.~Koltun, ``{CARLA: An
  Open Urban Driving Simulator},'' in \emph{Proc. 1st Ann. Conf. Robot Learn.},
  ser. Proc. of Mach. Learn. Research, {S. Levine et al.}, Ed., vol.~78.\hskip
  1em plus 0.5em minus 0.4em\relax PMLR, 13--15 Nov 2017, pp. 1--16.

\bibitem{melo2019simcomp}
M.~S.~P. de~Melo, J.~G. da~Silva~Neto, P.~J.~L. da~Silva, J.~M. X.~N. Teixeira,
  and V.~Teichrieb, ``{Analysis and Comparison of Robotics 3D Simulators},'' in
  \emph{Sympos. Virtual and Augmented Reality (SVR)}, 2019, pp. 242--251.

\bibitem{monteiro2019simulating}
F.~F. Monteiro, A.~L.~B. Vieira, J.~M. X.~N. Teixeira, V.~Teichrieb,
  \emph{et~al.}, ``{Simulating real robots in virtual environments using
  NVIDIA’s Isaac SDK},'' in \emph{Anais Estendidos do XXI Simp{\'o}sio de
  Realidade Virtual e Aumentada}.\hskip 1em plus 0.5em minus 0.4em\relax SBC,
  2019, pp. 47--48.

\bibitem{kamann2018auto}
A.~Kamann, P.~Held, F.~Perras, P.~Zaumseil, T.~Brandmeier, and U.~T. Schwarz,
  ``{Automotive Radar Multipath Propagation in Uncertain Environments},'' in
  \emph{Int. Conf. Intell. Transp. Syst. (ITSC)}.\hskip 1em plus 0.5em minus
  0.4em\relax IEEE, 2018, pp. 859--864.

\bibitem{holtzman1978}
J.~C. Holtzman, V.~S. Frost, J.~L. Abbott, and V.~H. Kaupp, ``{Radar Image
  Simulation},'' \emph{Trans. Geosc. Electron.}, vol.~16, no.~4, pp. 296--303,
  1978.

\bibitem{scheiblhofer2008}
S.~Scheiblhofer, M.~Treml, S.~Schuster, R.~Feger, and A.~Stelzer, ``{A
  versatile FMCW Radar System Simulator for Millimeter-Wave Applications},'' in
  \emph{Eur. Microwave Conf. (EuMC)}.\hskip 1em plus 0.5em minus 0.4em\relax
  IEEE, 2008, pp. 1604--1607.

\bibitem{dudek2010}
M.~Dudek, R.~Wahl, D.~Kissinger, R.~Weigel, and G.~Fischer, ``{Millimeter Wave
  FMCW Radar System Simulations including a 3D Ray Tracing Channel
  Simulator},'' in \emph{Asia-Pac. Microwave Conf. (APMC)}.\hskip 1em plus
  0.5em minus 0.4em\relax IEEE, 2010, pp. 1665--1668.

\bibitem{wheeler2017}
T.~A. Wheeler, M.~Holder, H.~Winner, and M.~J. Kochenderfer, ``{Deep Stochastic
  Radar Models},'' in \emph{Intell. Vehicles Symposium. (IV)}.\hskip 1em plus
  0.5em minus 0.4em\relax IEEE, 2017, pp. 47--53.

\bibitem{tang2020}
T.~Y. Tang, D.~De~Martini, D.~Barnes, and P.~Newman, ``{RSL-Net: Localising in
  Satellite Images From a Radar on the Ground},'' \emph{{IEEE} Robot. Automat.
  Lett.}, vol.~5, no.~2, pp. 1087--1094, 2020.

\bibitem{weston2019}
R.~Weston, S.~Cen, P.~Newman, and I.~Posner, ``{Probably Unknown: Deep Inverse
  Sensor Modelling In Radar},'' in \emph{Int. Conf. Robot. Automat.
  (ICRA)}.\hskip 1em plus 0.5em minus 0.4em\relax IEEE, 2019, pp. 5446--5452.

\bibitem{weston2021}
R.~Weston, O.~P. Jones, and I.~Posner, ``{There and Back Again: Learning to
  Simulate Radar Data for Real-World Applications},'' in \emph{Int. Conf.
  Robot. Automat. (ICRA)}.\hskip 1em plus 0.5em minus 0.4em\relax IEEE, 2021,
  pp. 12\,809--12\,816.

\bibitem{hirsenkorn2017ray}
N.~Hirsenkorn, P.~Subkowski, T.~Hanke, A.~Schaermann, A.~Rauch, R.~Rasshofer,
  and E.~Biebl, ``{A Ray Launching Approach for Modeling an FMCW Radar
  System},'' in \emph{Int. Radar Sympos. (IRS)}.\hskip 1em plus 0.5em minus
  0.4em\relax IEEE, 2017, pp. 1--10.

\bibitem{wald2019ray}
S.~O. Wald and F.~Weinmann, ``{Ray Tracing for Range-Doppler Simulation of 77
  GHz Automotive Scenarios},'' in \emph{Eur. Conf. Antennas and Propagation
  (EuCAP)}.\hskip 1em plus 0.5em minus 0.4em\relax IEEE, 2019, pp. 1--4.

\bibitem{holder2019}
M.~Holder, C.~Linnhoff, P.~Rosenberger, and H.~Winner, ``{The Fourier Tracing
  Approach for Modeling Automotive Radar Sensors},'' in \emph{2019 20th
  International Radar Symposium (IRS)}, 2019, pp. 1--8.

\bibitem{yilmaz2020radar}
E.~Yilmaz, ``{Radar Sensor Plugin for Game Engine Based Autonomous Vehicle
  Simulators},'' Ph.D. dissertation, Harvard Univ., 2020.

\bibitem{schuessler2021}
C.~Schüßler, M.~Hoffmann, J.~Bräunig, I.~Ullmann, R.~Ebelt, and M.~Vossiek,
  ``{A Realistic Radar Ray Tracing Simulator for Large MIMO-Arrays in
  Automotive Environments},'' \emph{{IEEE} Microwave}, pp. 1--13, 09 2021.

\bibitem{optix10}
S.~G. Parker, J.~Bigler, A.~Dietrich, H.~Friedrich, J.~Hoberock, D.~Luebke,
  D.~McAllister, M.~McGuire, K.~Morley, A.~Robison, and M.~Stich, ``{OptiX: A
  General Purpose Ray Tracing Engine},'' \emph{ACM Trans. Graphics (TOG)},
  vol.~29, no.~4, jul 2010.

\bibitem{maier2018phong}
M.~Maier, V.~P. Makkapati, and M.~Horn, ``{Adapting Phong into a Simulation for
  Stimulation of Automotive Radar Sensors},'' in \emph{Int. Conf. Microwaves
  Intell. Mobility (ICMIM)}.\hskip 1em plus 0.5em minus 0.4em\relax IEEE, 2018,
  pp. 1--4.

\bibitem{cgpp-book}
J.~F. Hughes, A.~van Dam, M.~McGuire, D.~F. Sklar, J.~D. Foley, S.~K. Feiner,
  and K.~Akeley, \emph{Computer graphics: principles and practice (3rd
  ed.)}.\hskip 1em plus 0.5em minus 0.4em\relax Boston, MA, USA: Addison-Wesley
  Professional, 7 2013.

\bibitem{mock2023rmagine}
A.~Mock, T.~Wiemann, and J.~Hertzberg, ``{Rmagine: 3D Range Sensor Simulation
  in Polygonal Maps via Ray Tracing for Embedded Hardware on Mobile Robots},''
  in \emph{Int. Conf. Robot. Automat. (ICRA)}.\hskip 1em plus 0.5em minus
  0.4em\relax IEEE, 2023, pp. 9076--9082.

\bibitem{embree14}
I.~Wald, S.~Woop, C.~Benthin, G.~S. Johnson, and M.~Ernst, ``{Embree: A Kernel
  Framework for Efficient CPU Ray Tracing},'' \emph{ACM Trans. Graphics (TOG)},
  vol.~33, no.~4, jul 2014.

\bibitem{wiemann2018lvr}
T.~Wiemann, I.~Mitschke, A.~Mock, and J.~Hertzberg, ``{Surface Reconstruction
  from Arbitrarily Large Point Clouds},'' in \emph{Int. Conf. Robotic Comput.
  (IRC)}.\hskip 1em plus 0.5em minus 0.4em\relax IEEE, 2018, pp. 278--281.

\bibitem{mock2024micpl}
A.~Mock, S.~Pütz, T.~Wiemann, and J.~Hertzberg, ``{MICP-L: Mesh ICP for Robot
  Localization using Hardware-Accelerated Ray Casting},'' in \emph{Int. Conf.
  Intell. Robots Syst. (IROS)}.\hskip 1em plus 0.5em minus 0.4em\relax IEEE,
  2024.

\bibitem{cook1982reflectance}
R.~L. Cook and K.~E. Torrance, ``{A Reflectance Model for Computer Graphics},''
  \emph{ACM Trans. Graphics (TOG)}, vol.~1, no.~1, pp. 7--24, 1982.

\bibitem{walter2007microfacet}
B.~Walter, S.~R. Marschner, H.~Li, and K.~E. Torrance, ``{Microfacet Models for
  Refraction through Rough Surfaces},'' in \emph{Proc. 18th Eurographics Conf.
  Rendering Techniques}, 2007, pp. 195--206.

\bibitem{pan2024pin}
Y.~Pan, X.~Zhong, L.~Wiesmann, T.~Posewsky, J.~Behley, and C.~Stachniss,
  ``{PIN-SLAM: LiDAR SLAM Using a Point-Based Implicit Neural Representation
  for Achieving Global Map Consistency},'' \emph{IEEE Trans. Robotics},
  vol.~40, 2024.

\bibitem{igelbrink2019calib}
F.~Igelbrink, T.~Wiemann, S.~Pütz, and J.~Hertzberg, ``{Markerless Ad-Hoc
  Calibration of a Hyperspectral Camera and a 3D Laser Scanner},'' in
  \emph{Conf. Intell. Auton. Syst. (IAS)}.\hskip 1em plus 0.5em minus
  0.4em\relax IAS, 01 2019, pp. 748--759.

\bibitem{whang2004ssi}
Z.~Wang, A.~Bovik, H.~Sheikh, and E.~Simoncelli, ``{Image quality assessment:
  from error visibility to structural similarity},'' \emph{{IEEE} Trans. Image
  Processing}, vol.~13, no.~4, pp. 600--612, 2004.

\end{thebibliography}

\end{document}